\documentclass{article} 
\usepackage{iclr2024_conference,times}

\usepackage{amsmath,amsfonts,bm}

\def\eqref#1{equation~\ref{#1}}

\def\1{\bm{1}}

\DeclareMathAlphabet{\mathsfit}{\encodingdefault}{\sfdefault}{m}{sl}
\SetMathAlphabet{\mathsfit}{bold}{\encodingdefault}{\sfdefault}{bx}{n}

\usepackage{hyperref}
\usepackage{url}
\usepackage{multirow}
\usepackage{booktabs}

\usepackage{graphicx} 
\usepackage{colortbl}
\usepackage{adjustbox}
\usepackage{comment}
\usepackage{makecell}
\usepackage{xspace}

\title{\ourMethod: Self-Training Enables Video Instruction Tuning with Any Supervision}

\author{Orr Zohar$^{1\dag}$, Xiaohan Wang$^1$, Yonatan Bitton$^2$, Idan Szpektor$^2$ \& Serena Yeung-Levy$^{1\dag}$ \\
$^1$Stanford University, $^2$Google Research\\
$^\dag$\texttt{\{orrzohar,syyeung\}@stanford.edu} 
\\
Project page: \url{https://orrzohar.github.io/projects/video-star/}
}

\definecolor{sgreen}{RGB}{30, 150, 30} 
\definecolor{amber(sae/ece)}{rgb}{1.0, 0.49, 0.0}
\definecolor{myblue}{RGB}{10,0,250}
\definecolor{blue}{RGB}{230,230,255}

\newcommand{\ourMethod}[0]{Video-STaR\xspace}
\newcommand{\ourDataset}[0]{VSTaR-1M\xspace}

\newcommand{\base}[0]{Video-LLaVA$^+$}
\newcommand{\geminiBase}[0]{Vid-LLaVA$^{Gemini}$}

\iclrfinalcopy 

\begin{document}

\maketitle

\vspace{-0.05in}

\begin{abstract}
The performance of Large Vision Language Models (LVLMs) is dependent on the size and quality of their training datasets. 
Existing video instruction tuning datasets lack diversity as they are derived by prompting large language models with video captions to generate question-answer pairs, and are therefore mostly descriptive.
Meanwhile, many labeled video datasets with diverse labels and supervision exist - however, we find that their integration into LVLMs is non-trivial. 
Herein, we present \underline{Video} \underline{S}elf-\underline{T}raining with \underline{a}ugmented \underline{R}easoning (\ourMethod), the first video self-training approach. \ourMethod allows the utilization of \textit{any} labeled video dataset for video instruction tuning.
In \ourMethod, an LVLM cycles between instruction generation and finetuning, which we show (I) improves general video understanding and (II) adapts LVLMs to novel downstream tasks with existing supervision. 
During generation, an LVLM is prompted to propose an answer. The answers are then filtered only to those that contain the original video labels, and the LVLM is then re-trained on the generated dataset. 
By only training on generated answers that contain the correct video labels, \ourMethod utilizes these existing video labels as weak supervision for video instruction tuning.
Our results demonstrate that \ourMethod-enhanced LVLMs exhibit improved performance in (I) general video QA, where TempCompass performance improved by $10\%$, \textit{and} (II) on downstream tasks, where \ourMethod\ improved Kinetics700-QA  accuracy by $20\%$ and action quality assessment on FineDiving by $15\%$.
\vspace{-0.05in}
\end{abstract}

\section{Introduction}
\vspace{-0.05in}

The advent of large Vision-Language Models (LVLMs) marked a significant milestone in artificial intelligence. 
These models aim to create versatile systems capable of understanding and executing vision-and-language tasks aligned with human intentions.
Early advancements in LVLMs, as exemplified by works such as BLIP~\citep{BLIP, BLIP2} and LLaVA~\citep{llava,llava2}, have been driven by the dissemination of pre-trained large language models (LLMs) (e.g., LLaMA~\citep{llama, llama2}) and pre-trained vision/vision-language models (e.g., CLIP~\citep{CLIP}). LVLMs connect the two model types via visual-language alignment and instruction tuning.

\begin{figure}[t]
  \centering
    \resizebox{
\linewidth}{!}{
    \includegraphics{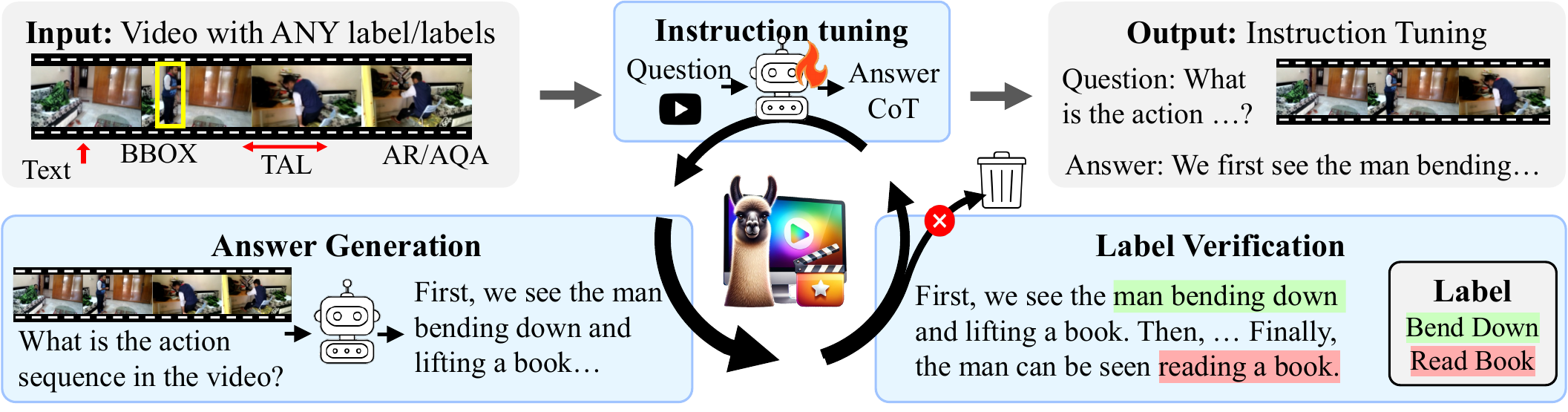}
    }\vspace{-0.1in}
\caption{\textbf{\ourMethod Overview.} 
\ourMethod\ can utilize \textit{any} labeled video dataset, including AR (Action Recognition), AQA (Action Quality Assessment), and TAL (Temporal Action Localization) -- from which it generates video instruction tuning data (video, question, answer triplets). 
Internally, \ourMethod\ cycles between: 
(I) \textbf{Answer Generation}, where an LVLM is prompted to generate candidate answers for the questions. 
(II) \textbf{Label Verification} where generated answers are filtered to only those that contain the video labels. 
And (III) \textbf{Instruction Tuning}, where a model is retrained on answers that pass verification. These cycles continue until performance plateaus. \vspace{-0.15in}
} 
\label{fig:poll}
\end{figure}

\citet{finetune2,llava2,prismaticVLM} demonstrated the importance of visual instruction tuning on the resulting LVLM's performance.
However, while much progress has been made in image-LVLMs, video-LVLMs still face challenges due to the relative difficulty in generating quality video instruction tuning datasets.
Even though videos contain more complex scene dynamics and temporal information, indicating a need for larger and more diverse training datasets compared to images, the largest video instruction dataset, VideoInstruct-$100K$ (VI-$100K$)~\citep{videoChatGpt}, comprises $100K$ video-text pairs but only $13K$ unique videos. This is small compared to image instruction datasets like LLaVA-$1.5$-Instruct~\citep{llava2}, which has $\sim665K$ image question pairs and $\sim350K$ unique images. Such limitation in video instruction tuning leads to performance saturation~\citep{videoChat,videoChatGpt, chatUniVi,VaQuitA,videoLlava}.

Furthermore, due to video instruction tuning dataset construction - mainly prompting large language models to produce question-answer pairs - these video datasets often degrade to simplistic questions, prompting for video captions  — $75$\% of VI-$100K$'s questions are of this type (see App. Fig.~\ref{sup:fig:video_instruction_quality}). 
Combining different sources of supervision has the potential to generate more diverse video instruction tuning datasets, enhancing video understanding. 
Such supervision exists, as the broader computer vision community has developed an extensive collection of video benchmarks tailored for diverse tasks such as action recognition~\citep{Kinetics700, UCF100}, action quality assessment~\citep{FineDiving, LOGO}, among others~\citep{ActivityNet,STARDataset,agqa, EgoExo4D, MOMA}.

Beyond improving overall LVLM performance, adapting LVLMs to novel or out-of-domain tasks is also crucial. While LVLMs have many novel and impactful applications, many remain out of reach — such as analyzing radiology images~\citep{gpt4v_bapp_med}, meteorological data~\citep{gpt4v_bapp}, traffic analysis~\citep{gpt4v_bapp1}, judging sporting events (e.g., gymnastics, Olympic diving), and assisting in surgical procedures, among others~\citep{gpt4v_bapp2,gpt4v_bapp_med2,gpt4v_bapp_med3}. These tasks require expert, in-domain knowledge that LVLMs lack, necessitating adaptation through instruction tuning. 
However, collecting video instruction tuning datasets is complex and requires extensive manual effort. For instance, training an `AI judge' to judge Olympic diving would traditionally involve collecting detailed expert critiques of each dive. On the other hand, these tasks often include auxiliary annotations that could be leveraged, such as surgical outcomes in medical procedures or judging scores in Olympic events.

To address these challenges, we take inspiration from LLM's capability for self-improvement (a.k.a self-training)~\citep{SelfImprove, STaR}, which involves training a model on its generated data and filtering to exclude low-quality outputs. 
A model's performance is improved by cycling between generation, filtering, and training.
For example,~\cite{STaR} introduced Self-Taught Reasoners which generate chain-of-thought by prompting an LLM to generate answers and rationalize responses, retaining only correctly answered questions for further training. 
Herein, we explore self-training in LVLMs and introduce \underline{Video} \underline{S}elf-\underline{T}raining with \underline{a}ugmented \underline{R}easoning (\ourMethod, see Fig.~\ref{fig:poll}).
\ourMethod enables the incorporation of \emph{any} 
labeled video dataset of any format or task by prompting an LVLM with the video and a question
to generate answers (Fig.~\ref{fig:method}, 2.1) containing the video content's label. 
If the model cannot correctly answer the question, we provide the video label and ask it to rationalize it (Fig.~\ref{fig:method}, 2.2). 
We then utilize the labels as weak supervision, rejecting answers that do not contain the correct label (Fig.~\ref{fig:method}, 2.3).
By facilitating the use of any supervision for video instruction tuning, \ourMethod enables the creation of diverse datasets.

Our experimental setup initializes \ourMethod\ with Video-LLaVA~\citep{videoLlava}, focusing on assessing its impact on video question-answering (VQA) performance. 
After a few \ourMethod training cycles, we compare the performance of \ourMethod to other state-of-the-art LVLMs and strong baselines to gauge the effectiveness of the \ourMethod framework. 
Our findings demonstrate notable enhancements in accuracy and reasoning capabilities, highlighting \ourMethod's role in overcoming the constraints posed by conventional video instruction tuning datasets.
We show that the integration of \ourMethod not only boosts Video-LLaVA's performance on standard zero-shot VQA benchmarks but also significantly improves its adaptability to various downstream video understanding tasks. This underscores \ourMethod's capacity to advance LVLM training while improving overall performance and versatility.

\noindent\textbf{Our contributions can be summarized as follows:}
\vspace{-0.015in}
\begin{enumerate}  
    \item We introduce \underline{Video} \underline{S}elf-\underline{T}raining with \underline{a}ugmented \underline{R}easoning (\ourMethod), the first video self-training Large Video-Language Model method. Using self-training, \ourMethod enables the use of \textit{any} labeled video dataset for video instruction tuning. 
        
    \item \ourMethod improves zero-shot video question answering performance on various benchmark datasets, as evidenced by increased accuracy on TempCompass~\citep{tempcompass} from $45.7$ to $50.3$ ($+ 10\%$) and on MSVD-QA~\citep{videoChatGpt} from $69.7$ to $71.3$ ($+ 2.3\%$). 
    
    \item We demonstrate that \ourMethod can adapt LVLMs to diverse video tasks,  notably enhancing action quality assessment on FineDiving~\citep{FineDiving}, where accuracy rose from $17.6$ to $20.2$ ($+ 15 \%$) and in action recognition on Kinetics700~\citep{Kinetics700}, where accuracy increased from $50.0$ to $59.9$ ($+ 20\%$).

    \item Utilizing \ourMethod, we create a large, $1M$ video instruction tuning dataset - \ourDataset, sourced from diverse datasets and tasks, and show that it benefits LVLM performance.     
\end{enumerate}

\vspace{-0.05in}
\section{Video Self-Training with augmented Reasoning (\ourMethod)\label{sec:meth}}
\vspace{-0.05in}

Given a dataset of videos $v$ and their corresponding labels $l: \mathcal{D}= \{(v_i, l_i)\}_{i=1}^d$, \ourMethod's objective is to create question $q$ answer $a$ pairs to instruction-tune the pre-trained model $M$ on the dataset $\hat{\mathcal{D}}= \{(v_i, q_i, a_i)\}_{i=1}^{d_f}$, producing the instruction-tuned model $\hat{M}$. Note that videos need not be from the same task, and may contain multiple labels.
We start by prompting a large language model with a task description $T$ and video labels $L$ to generate candidate questions $q$:
\begin{align}
    Y_{T, L} &= \parbox[t]{\dimexpr\linewidth-3cm}{
    A video is labeled \{L\} for the task of \{T\}.
    What questions could you ask someone about the video that should contain the video labels in the response?
    }\nonumber
\end{align}
\ourMethod\ performs generation-training cycles, where in cycle $i$ the instruction-tuned model $\hat{M}^{i\star}$ is produced, while the instruction-tuned model from the previous cycle $\hat{M}^{(i-1)\star}$ is utilized for training data generation. We initialize the process with $\hat{M}^{0\star}$, an existing instruction-tuned model.

\begin{figure*}[t]
  \centering
    \resizebox{\linewidth}{!}{
    \includegraphics{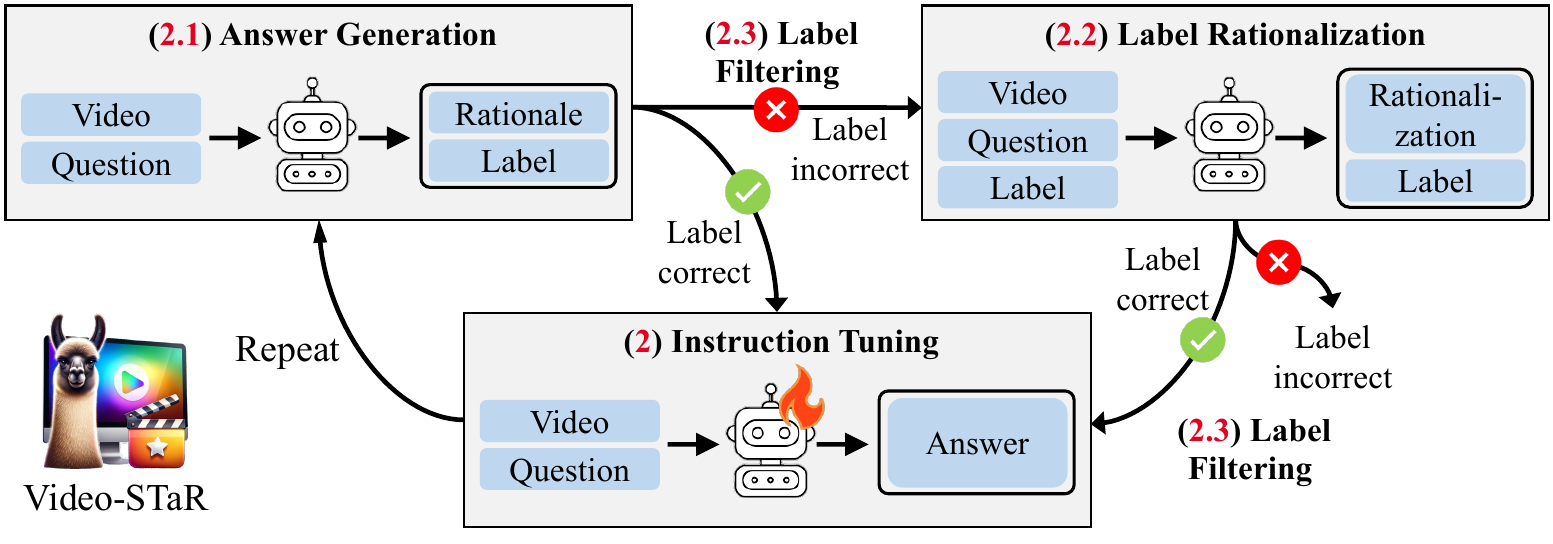}
    }
    \vspace{-0.15in}
  \caption{
\textbf{Video Self-Training with augmented Reasoning.} 
\textbf{(\ref{sec:meth:rationale})} We initialize by prompting an LVLM to generate an answer for a particular video. 
\textbf{(\ref{sec:meth:filter})} We then filter the generated answers to those only containing the original video labels.
\textbf{(\ref{sec:meth:rationalization})} The videos whose generated answer did not contain the ground-truth labels are then sent to label rationalization, where given the video, question, and label - the model is expected to rationalize the label. \textbf{(\ref{sec:meth:filter})} The generated answers are filtered again to those only containing the ground-truth labels, and 
\textbf{(\ref{sec:meth})} the LVLM is instruction-tuned from the pre-trained checkpoint on the resulting dataset. The cycle is then repeated.
  }
\label{fig:method}
\end{figure*}

To prepare the training data in cycle $i$, answers are generated either directly via \textit{Answer Generation} or through backward rationalization via \textit{Label Rationalization}. In \textit{Answer Generation}, $\hat{M}^{(i-1)\star}$ is prompted with questions (Sec.~\ref{sec:meth:rationale}). Candidate answers are then filtered using the original video labels (Sec.~\ref{sec:meth:filter}). Videos rejected during direct Answer Generation are rationalized, where $\hat{M}^{(i-1)\star}$ is provided both a video $v_i$ and labels $l_i$, and then prompted with the question again (Sec.~\ref{sec:meth:rationalization}). Candidate answers are filtered again, creating the instruction tuning dataset in cycle $i$, $\hat{\mathcal{D}_i}$ of size $d_i$. A pre-trained model $M$ is then finetuned on $\mathcal{D}_i$, producing $\hat{M}^{i\star}$. The next cycle generates data using $\hat{M}^{i\star}$, until the performance plateaus (see Fig.~\ref{fig:method}).

\vspace{-0.05in}
\subsection{Answer Generation}
\vspace{-0.05in}
\label{sec:meth:rationale}
Each \ourMethod cycle begins in direct Answer Generation. 
In this phase, $\hat{M}^{(i-1)\star}$ is prompted with the video-question pair to provide an answer along with a detailed rationale: 
\begin{align}
    Y_{Q} &= \parbox[t]{\dimexpr\linewidth-3cm}{
    Question: \{Q\}.
    Rationalize your answer step-by-step; how can one arrive at this conclusion?
    }\nonumber
\end{align}
When prompted with the question $q_i$ on a particular video, $\hat{M}^{(i-1)\star}$ is expected to generate an answer $a_i$ that contains the label $\hat{l}_i$ and the rationale $r_i$ ($a_i = r_i \cup \hat{l}_i$, see Fig.~\ref{fig:method}).
We observe that answers containing the correct labels are of higher quality and suffer less from hallucination. 
Therefore, we filter the generated answers to include only those that contain the correct label $(\hat{l}_i = l_i)$ utilizing a verifier (Sec.~\ref{sec:meth:filter}). 
For an example of Answer Generation, see Fig.~\ref{fig:data_gen_qual}.

\vspace{-0.05in}
\subsection{Label Rationalization}
\vspace{-0.05in}
\label{sec:meth:rationalization}
Answer generation has two main drawbacks: (i) in some applications, especially on challenging/out-of-domain tasks, initial answer generation yield is low, resulting in almost no training samples after filtering (e.g., FineDiving, see Fig.~\ref{fig:data_gen_qual}); (ii) improvement plateaus as the model fails to solve new problems in the training set, and it is only trained on examples it answers correctly. 

Inspired by~\citet{STaR}, for videos whose $\hat{M}^{(i-1)\star}$ generated answers did not contain the ground-truth labels, we introduce \textit{label rationalization} as part of \ourMethod. Concretely, we provide $\hat{M}^{(i-1)\star}$ the video, question, and video label and instruct the model to rationalize the label:
\begin{align}
    Y_{Q, L} &= \parbox[t]{\dimexpr\linewidth-3cm}{
    Question: \{Q\}. 
    Answer: \{L\}.\\
    Can you rationalize the answer step-by-step? How can one arrive at this conclusion?
    }
    \nonumber
\end{align}

Given the correct label, the model can reason backward and more easily generate a rationale leading to the correct answer. However, label rationalization is more prone to hallucinations, so we prefer direct answer generation and use rationalizations only if answer generation fails. For an example of Label Rationalization, see Fig.~\ref{fig:data_gen_qual}, top. The generated answers are then filtered again, keeping only those that contain the correct label $(\hat{l}_i = l_i)$ utilizing a verifier (Sec.~\ref{sec:meth:filter}). 
Label Rationalization is only utilized in training cycles; only direct answers are used to produce the final model $\hat{M}^{\star}$.

\subsection{Label Verification}
\label{sec:meth:filter}
\ourMethod\ aims to utilize the labels as weak supervision in instruction tuning data generation. Gold labels are a grounding aspect of our datasets and represent some ground-truth knowledge. Therefore, we assume that answers that contain the ground-truth labels in their responses are higher quality than those that don't. 
While we would like to validate the existence of the different labels in the generated text, this can be non-trivial.

To this end, we introduce the Parser-Verifier. The \textit{Parser}, $P$ extracts the predicted labels from the generated text ($\hat{l_i}=P(a_i)$), using a mixture of named entity recognition and Regex. Regex is used to identify easily identifiable string patterns, such as bounding boxes and time ranges, while named entity recognition is used for more nuanced entities, such as timestamps. 
The \textit{Verfier}, $V$ compares the extracted labels with the gold ones using the appropriate metrics ($V(l_i, \hat{l_i})\rightarrow \mathbb{R}$). For example, IoU for bounding boxes/temporal action localization, and BERT~\citep{BERT} embedding similarity for sentence ordering. Each video has between 1-3 associated labels. To be classified as correct, the predicted labels must be within a 5\% margin of error from the gold.

\begin{table}[t]
  \setlength\tabcolsep{4.2mm}
  \label{tab:dataset_sources}
  \centering
  \adjustbox{width=\textwidth}{
  \begin{tabular}{l|cccl}
    \toprule
    \textbf{Source} &\textbf{Videos}  &  \textbf{Labels} &  \textbf{Avg. Dur.} &  \textbf{Source Task}  \\

    \midrule
    Kinetics700~(\citeyear{Kinetics700})& 650K & 700  & 5.1 & Action Recognition\\[0.2cm]
    \multirow{3}{*}{\makecell[l]{STAR\\benchmark\\ (\citeyear{STARDataset})}} & \multirow{3}{*}{22K}   & \multirow{3}{*}{207} & \multirow{3}{*}{28.6} & Video Reasoning, Temporal\\
     &    &  & &   Action Localization,\\
        &    &  &  &  Bounding Box\\[0.2cm]

    \multirow{2}{*}{\makecell[l]{FineDiving\\ (\citeyear{FineDiving})}} & \multirow{2}{*}{3K}  & \multirow{2}{*}{1065} & \multirow{2}{*}{3.2} & Action Quality Assessment,\\
     &   & & & Action Sequence\\
    \bottomrule
  \end{tabular} 
  }
  \caption{\textbf{Source Dataset Summary.} Video datasets used as source datasets for instruction tuning data generation, their source tasks, and the number of unique labels.}

\end{table}

\section{\ourMethod Generated Dataset - \ourDataset}
\label{sec:DataGen}
In this section, we detail the different source datasets utilized in our study (Sec.~\ref{sec:source_data}) and analyze the generated \ourMethod\ Dataset, \ourDataset (Sec.~\ref{sec:gen_data}).

 \begin{figure*}[t]
  \centering
    \resizebox{\linewidth}{!}{
    \includegraphics{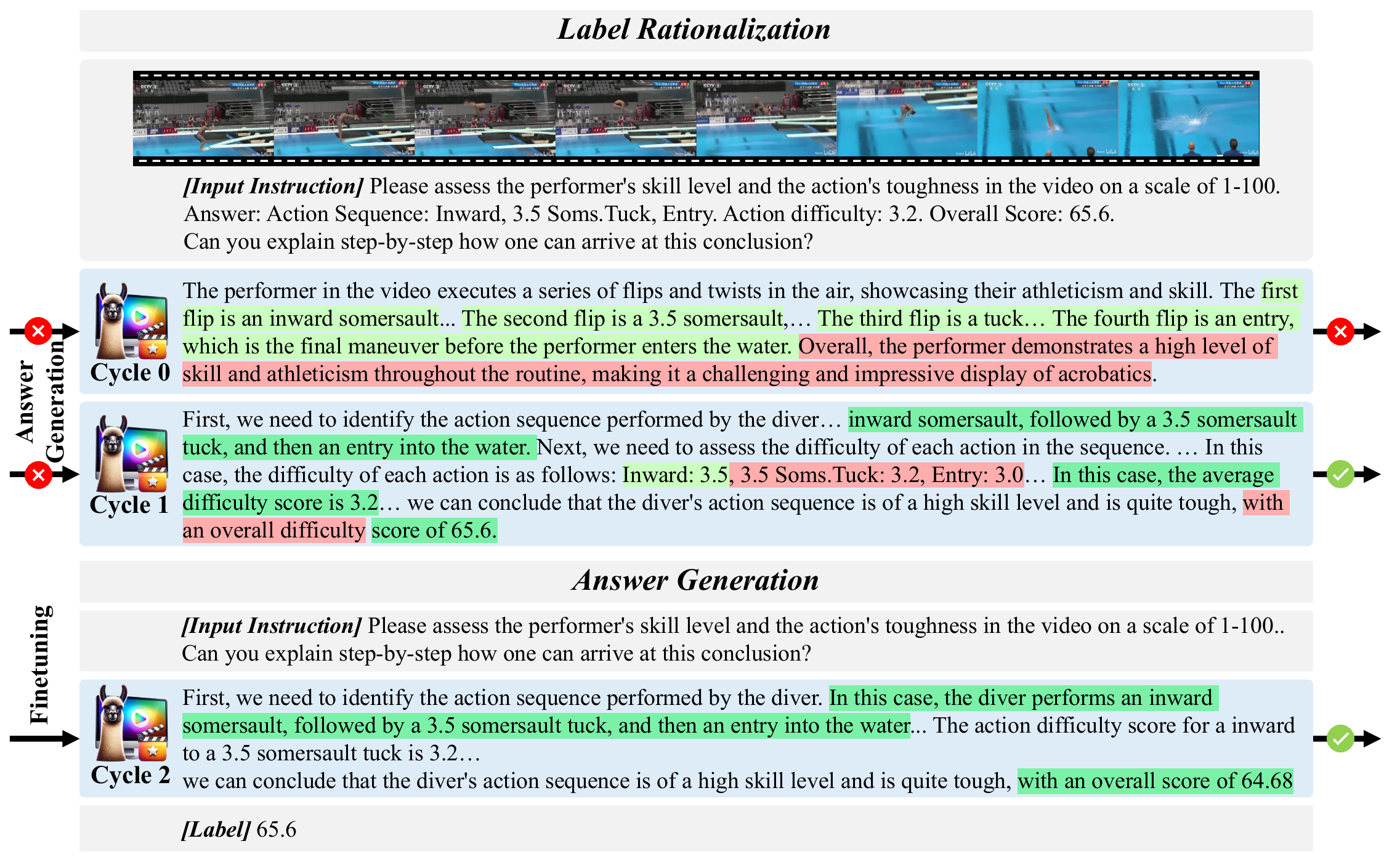}
    }
    \vspace{-0.15in}
  \caption{
  \textbf{Qualitative Improvement of Data Generation over Cycles on FineDiving.} We initialize the model with Video-LLaVA (Cycle 0), where the model cannot generate an answer ({\color{red}$\rightarrow\vert
\times$}) or rationalize the label correctly ({\color{red}$\vert\rightarrow
\times$}). In the second cycle (Cycle 1), the model still cannot generate an answer ({\color{red}$\rightarrow\vert
\times$}) but can rationalize the video label ({\color{green}$\checkmark\vert\rightarrow
$}), which is selected for instruction tuning. Finally, in the third cycle (Cycle 2), the model directly generates a correct answer ({\color{green}$\checkmark\vert\rightarrow
$}), which is selected for visual instruction tuning. We highlight in \textcolor{sgreen}{green} correct answers, in \textcolor{red}{red} wrong answers, and in \textcolor{amber(sae/ece)}{yellow} - hallucinations. 
  }\vspace{-0.15in}
\label{fig:data_gen_qual}
\end{figure*}

\subsection{Source Datasets}\label{sec:source_data}
In selecting source datasets, we selected datasets that contain diverse video content and label types, please see Tab.~\ref{tab:dataset_sources}. 
These include Kinetics700~\citep{Kinetics700}, which has action recognition annotations and is particularly large and diverse. FineDiving~\citep{FineDiving} is an action quality assessment dataset of Olympic diving events and has both an overall score and action sequence annotations. Finally, STAR-benchmark~\citep{STARDataset}, a video reasoning dataset,
also contains bounding box and temporal action localization annotations. Tab.~\ref{tab:dataset_sources} contains the relevant dataset statistics, e.g., the number of videos and labels per dataset.

\subsection{Generated Dataset Analysis}
\label{sec:gen_data}

\begin{figure*}[t]
  \centering
    \resizebox{\linewidth}{!}{
    \includegraphics{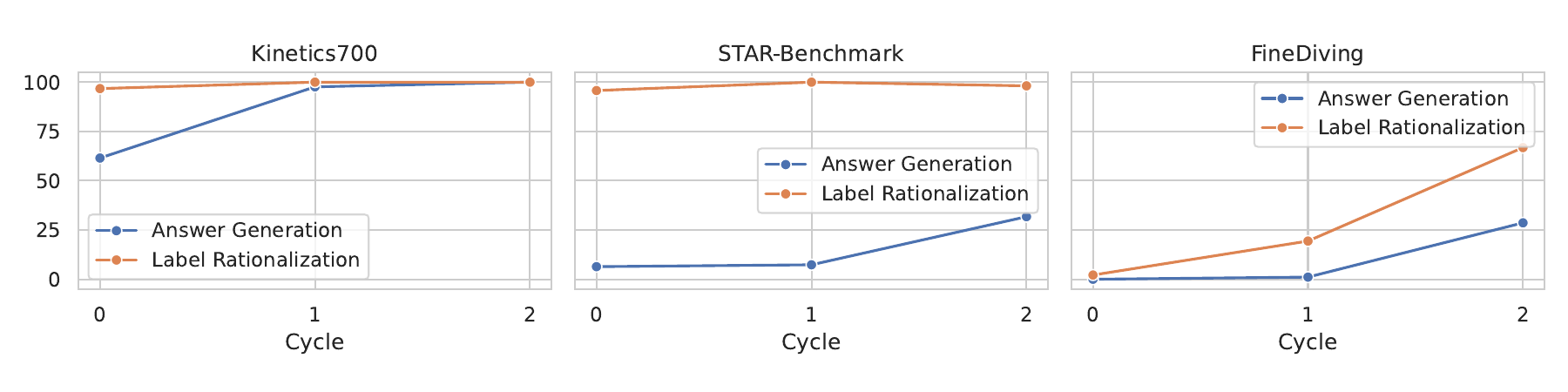}
    }
    \vspace{-0.38in}
  \caption{
  \textbf{Dataset Yield vs. Cycles.} Percentage of the videos converted to instruction tuning by the Answer Generation and Label Rationalization per dataset. As can be seen, on difficult datasets, such as FineDiving, no videos are converted by Answer Generation in the first cycle. By utilizing Label Rationalization, the model is able to improve to eventually generate answers correctly.  
  }\vspace{-0.1in}
\label{fig:progress}
\end{figure*}

\paragraph{Quantitative Analysis.}
Through the application of \ourMethod, significant dataset augmentation was achieved over two cycles of, illustrated in Fig.~\ref{fig:progress}. 
This figure displays the Answer Generation and Label Rationalization yield across the datasets source.
Notably, the initial application of Video-LLaVA on datasets like Kinetics700 and STAR-Benchmark showed significant Answer Generation success rates. 
However, the FineDiving dataset presented a notable challenge, with Answer Generation having no answers generated directly, underscoring the complexity of the dataset and the critical role of Label Rationalization. By the end of the second cycle, a substantial number of high-quality instances had been produced, showcasing both the effectiveness of \ourMethod\ in converting labeled video datasets into video instruction tuning dataset, as evidenced in Fig.~\ref{fig:progress}.

\paragraph{Qualitative Analysis.} See Tab.~\ref{tab:dataset_examples} for examples of generated question-answer pairs. 
From Kinetics700, we extracted an instance showcasing a video labeled `smashing'. Video-STaR correctly identified a more fine-grained label, `chopping wood'.
In the FineDiving dataset, a clip depicting a complex dive was accompanied by the question `On a scale from 1-100$\cdots$' The model's output text provided a breakdown of the dive's components, leading to a score (label), as would be desired from an LVLM visual assistant. 
Finally, in the STAR benchmark, questions are already provided; therefore, we utilized them directly.

\begin{table}[t]
\centering

\label{tab:dataset_examples}
\resizebox{\textwidth}{!}{ 
\begin{tabular}{lccc}
\toprule
 & \textbf{Kinetics700} & \textbf{STAR-Benchmark} & \textbf{FineDiving} \\
\midrule
Video & \includegraphics[width=0.4\textwidth]{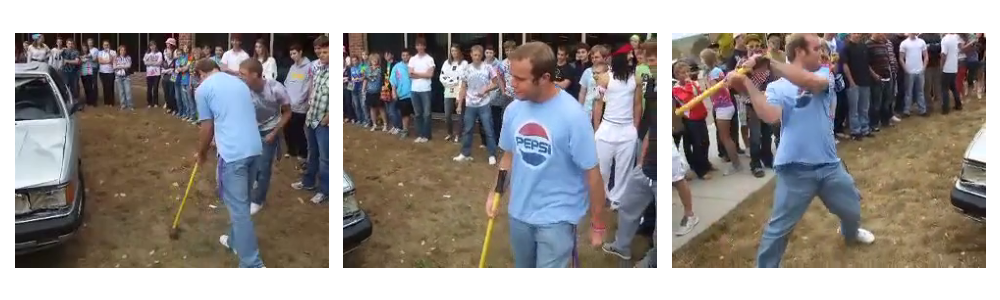} & \includegraphics[width=0.4\textwidth]{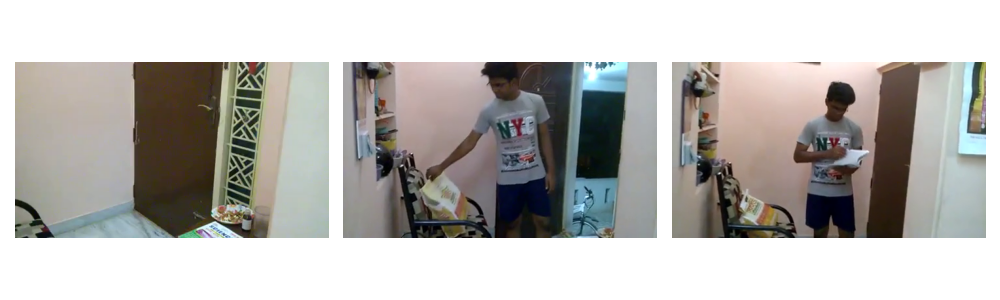} & \includegraphics[width=0.4\textwidth]{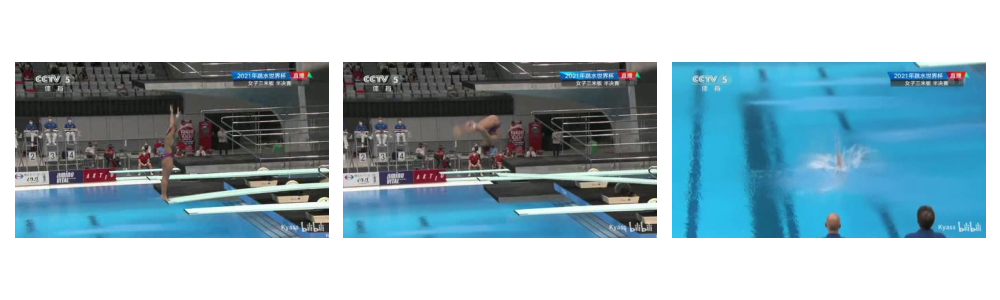} \\
\midrule
\multirow{1}{*}{\begin{minipage}{1.5cm} Question\end{minipage}} & \multicolumn{1}{m{5cm}}{What type of activity is being performed in this video?} & \multicolumn{1}{m{5cm}}{Which object was put down by the person?} & \multicolumn{1}{m{5cm}}{On a scale from 1-100, how would you evaluate the execution of the actions in the video?} \\
\midrule
\multirow{2}{*}{\begin{minipage}{1.5cm} Generated \\ Answer\end{minipage}} & \multicolumn{1}{m{5cm}}{In the video, a man is seen holding a large axe and swinging it around while a crowd watches. He then proceeds to hit a block of wood with the axe. Based on these observations, it can be concluded that the activity is chopping wood.} & \multicolumn{1}{m{5cm}}{In the video, the person is seen picking up a bag and putting it down on the floor. After that, he picks up a book and puts it down on the floor. Based on this information, we can conclude that the person put down the bag and picked up the book.} & \multicolumn{1}{m{5cm}}{First, we identify the diver's action sequence: inward somersault $\cdots$ entry into the water. Evaluating each movement's difficulty, we assign $\cdots$ inward somersault: 2.5,  $\cdots$. By averaging these scores $\cdots$ the overall score is 63.0 out of 100.} \\
\midrule
Label & Smashing & The bag. & Overall Score 63.0 \\
\bottomrule
\end{tabular}
}
\vspace{-0.05in}
\caption{\textbf{Examples of Generated Data.} Examples of the video, question, \ourMethod generated answer, and ground-truth label from each source dataset.\vspace{-0.15in}}
\end{table}

 In~\ref{fig:data_gen_qual}, we show the qualitative improvement of the generated data over \ourMethod\ cycles. 
 In the first cycle (Cycle 0), Video-LLaVA failed at Answer Generation and Label Rationalization. After one \ourMethod\ cycle (Cycle 1), \ourMethod\ still failed at Answer Generation but succeeded in Label Rationalization. After the final \ourMethod\ cycle (Cycle 2), \ourMethod\ managed to generate the answer without requiring the label via Answer Generation.

\section{Experiments}
We experimented with \ourMethod\ and evaluated its enhanced video understanding capabilities. In Sec.~\ref{sec:res:adapt}, we evaluate how \ourMethod\ adapts Large Vision-Language Models (LVLMs) to the source datasets and how these capabilities are transferred zero-shot to similar benchmarks. In Sec.~\ref{sec:res:zs}, we evaluate the video question-answering capabilities on video benchmark datasets. 

\subsection{Experimental Setting}

\paragraph{Implementation Details.} We initialize from the Video-LLaVA~\citep{videoLlava} model, which utilizes the Vicuna-7B v1.5~\citep{Vicuna}. We ran three \ourMethod\ cycles, and each cycle was initialized with the pre-trained Video-LLaVA weights.
We train for one epoch using a $128$ batch size, AdamW optimizer, and a cosine learning rate schedule. The learning rate is $2e-5$ with a $0.03$ warmup ratio. 
In combination with the generated \ourMethod instruction tuning dataset, we additionally utilized the VideoInstruct-$100K$~\citep{videoChatGpt} and the LLaVA v$1.5$ instruction tuning datasets~\citep{llava2}.
Additional details are available in the appendix.

\paragraph{Baselines.} 
Besides comparing to Video-LLaVA, we also wanted to evaluate the effect of utilizing additional data and naively adapting the source datasets. Therefore, we utilized simple templates to generate question-answer pairs from the video labels and trained Video-LLaVA on the resulting dataset. We will reference this baseline as \base. 
Another baseline for adapting Large Vision-Language Models to novel tasks is model distillation, where a stronger video model - in this work, Gemini $1.5$ pro-vision - is utilized to label/annotate a small set of videos ($500$ from each dataset) and used to finetune the models. We will reference this baseline as \geminiBase.

\paragraph{Evaluation Details.} We evaluate on the following benchmarks; the Zero-shot question-answer (QA) benchmarks: MSVD-QA, MSRVTT-QA, TGIF-QA, and ActivityNet-QA~\citep{MSVD,MSRVTT,TGIF,ActivityNet}. TempCompass~\citep{tempcompass}, a multiple-choice fine-grained QA benchmark.
Adapted task performance is evaluated by converting source datasets using simple templates and applying the same evaluation protocol as Maaz et al.~\citep{videoChatGpt}, producing Kinetics700-QA, STAR-benchmark-QA, and FineDiving-QA.
 This protocol reports two metrics: accuracy (the percentage of correctly answered questions) and the average score (where ChatGPT rates each response on a scale of 1-5 and calculates the mean of these scores).
All evaluations utilize the same GPT model~\citep{freeva} (``gpt-3.5-turbo'') to ensure consistent comparisons.
Due to cost considerations, 1000 videos were randomly selected from each dataset for Gemini evaluation. The reported values are used on ActivitlyNet-QA.

\begin{table*}[t]

\centering 
\resizebox{\textwidth}{!}{%
\begin{tabular}{@{}l|cc|cc|cc|c|cccc|c@{}}
\toprule 
& \multicolumn{2}{c|}{\textbf{Action}} & \multicolumn{2}{c|}{\textbf{Direction}} & \multicolumn{2}{c|}{\textbf{Speed}} & \textbf{Event} & \multicolumn{4}{c|}{\textbf{Attribute Change}} & \textbf{Avg.} \\ 

& {Fine} & {Coarse} & {Obj.} & {Cam.} & {Abs.} & {Rel.} & {Order} & {Color} & {Size} & {Both} & {Other} &  \\ 
\midrule
Random & 39.7 & 40.1 & 39.8 & 39.0 & 40.8 & 42.0 & 41.5 & 40.4 & 39.9 & 38.9 & 39.4 & 40.5 \\
mPLUG-Owl~(\citeyear{mplugowl}) & 48.8 & 66.1 & 38.7 & 36.8 & 42.2 & 38.4 & 42.0 & 41.7 & 44.7 & 41.9 & 39.9 & 44.4 \\
Video-LLaVA~(\citeyear{videoLlava}) & 63.4 & 93.5 & 36.1 & 34.8 & 42.7 & 26.5 & 39.1 & 52.6 & 37.1 & 43.3 & 33.3 & 45.7 \\
\base & 62.1 & 93.0 & 35.0 & 32.6 & 41.1 & 38.7 & 36.4 & 59.0 & 40.2 & 36.7 & 44.4 & 47.2 \\
\geminiBase &  30.7 & 30.1 & 37.8 & 40.0 & 41.8 & 42.4 & 21.5 & 50.4 & 49.9 & 38.0 & 37.4 & 38.2\\
 \rowcolor{blue} \ourMethod & 68.6 & 94.1 & 35.8 & 38.0 & 38.7 & 37.6 & 37.1 & 53.8 & 48.5 & 45.0 & 55.6 & 50.3$(+10\%)$\\
 \hline
 Gemini-1.5~(\citeyear{gemini}) & 94.8 & 98.4 & 43.6 & 42.4 & 65.3 & 48.7 & 55.6 & 79.5 & 59.8 & 70.0 & 66.7 & 66.0 \\

\bottomrule 
\end{tabular}%
}
\vspace{-0.05in}
\caption{\textbf{Comparison with state-of-the-art methods on TempCompass.} TempCompass~\citep{tempcompass} assesses the temporal understanding capabilities of video language models across five dimensions \ourMethod\ improves Video-LLaVA performance on TempCompass by $10\%$.\label{tab:tempcompass}}

\end{table*}

\begin{table}[tb]
  \setlength\tabcolsep{0.5mm}
  \centering
  \adjustbox{width=\textwidth}{
  \begin{tabular}{lc|cc|cc|cc|cc}
    \toprule
    \multirow{2}{*}{\textbf{Methods}} & \textbf{Dataset}& \multicolumn{2}{c|}{\textbf{MSVD-QA}} & \multicolumn{2}{c|}{\textbf{MSRVTT-QA}} & \multicolumn{2}{c|}{\textbf{TGIF-QA}} & \multicolumn{2}{c}{\textbf{ActivityNet-QA}} \\
      & \textbf{size} & Accuracy & Score & Accuracy & Score & Accuracy & Score & Accuracy & Score \\
     \midrule
    VideoChat~(\citeyear{videoChat}) & 4K & 56.3 & 2.8 & 45.0 & 2.5 & 34.4 & 2.3 & - & 2.2 \\
    Video-LLaMA~(\citeyear{videollama}) & 4K & 51.6 & 2.5 & 29.6 & 1.8 & - & - & 12.4 & 1.1 \\

    Video-ChatGPT~(\citeyear{videoChatGpt})  & 100K & 64.9 & 3.3 & 49.3 & 2.8 & \textbf{51.4} & \underline{3.0} & 35.2 & 2.7 \\

    Video-LLaVA$^*$~(\citeyear{videoLlava})  & 100K & \underline{69.7} & \underline{3.9} & \underline{57.4} & \textbf{3.5} & 46.5 & \textbf{3.3} & \textbf{43.2} & \textbf{3.4} \\
    
     \base\  & 650K & 67.8 & 3.8 & 56.0 & \underline{3.4} & 46.5 & \textbf{3.3} & \underline{42.2} & \underline{3.3} \\

     \geminiBase\  & 2k & 67.2 & 3.9 & 55.9 & 3.5 & 44.5 & 3.2 & 39.6 & 3,3 \\

    \rowcolor{blue} \ourMethod & 550K & \textbf{71.3} & \textbf{4.0} &\textbf{ 58.2} & \textbf{3.5} & \underline{47.3} & \textbf{3.3} & \textbf{43.2} & \underline{3.3} \\

    \midrule
    Gemini-1.5-pro~(\citeyear{gemini}) & - & 71.6 & \underline{3.9} & 52.6 & 3.2 & 45.0 & 3.1 & 56.7 &  -  \\

    \bottomrule
  \end{tabular} 
  }
  \vspace{-0.05in}
  \caption{\textbf{Zero-shot Video QA benchmarks}. As can be seen, many models are approaching Gemini performance - indicating that LVLMs may be operating near the noise level on these benchmarks.  \label{tab:video_qa}}

\end{table}

\subsection{Quantitative Evaluation on Zero-Shot Benchmarks} \label{sec:res:zs}
To evaluate \ourMethod's effect on general video question answering, we evaluated its effect on Video-LLaVA's performance on TempCompass, see Tab.~\ref{tab:tempcompass}. 
On TempCompass, \ourMethod\ outperformed Video-LLaVA across the board-- by $\sim10\%$. 
To see if this performance boost is simply a factor of training on a larger dataset, we also evaluated \base. \base was trained on even a larger video dataset by naively utilizing video labels, and yields a more modest improvement of $3\%$, showing the utility of \ourMethod.
TempCompass is also a fine-grained dataset that would be sensitive to hallucinations, indicating that \ourMethod\ is not more prone to hallucinations compared to existing methods. Gemini 1.5 pro scored an impressive $66.0$ on TempCompass, showing there is still much room for improvement on this benchmark.

We then continued and evaluated \ourMethod's effect on zero-shot video QA performance on the MSVD-QA, MSRVTT-QA, TGIF-QA and ActivityNet-QA benchmarks. As can be seen in Tab.~\ref{tab:video_qa}, \ourMethod\ achieves performance improvements where, for instance, on the MSVD-QA dataset, \ourMethod attains the highest accuracy of $71.3$\% vs Video-LLaVA's 69.7. 
On MSRVTT-QA, \ourMethod leads with an accuracy of $58.2$\% and maintains a competitive edge in other datasets like TGIF-QA and ActivityNet-QA. 
Seeing the relatively small performance gains compared to TempCompass, we additionally evaluated Gemini $1.5$ pro-vision on $1000$ video subsets of each dataset and found that its performance is on par with existing open-source models.
We believe this shows that we are near the `noise' limit of these benchmarks. 
Our qualitative analysis indicated that many of the questions selected as `wrong' are actually due to the benchmark design—overly general questions with multiple correct answers. Concurrent work~\citep{freeva} has similarly concluded that the ChatGPT-$3.5$ version utilized in evaluation can lead to variations of $\pm 10$ in accuracy.

\begin{table}[tb]
  \setlength\tabcolsep{1mm}
  \centering
  \adjustbox{width=\textwidth}{
    \begin{tabular}{l|cc|cc|cc}
      \toprule
      \multirow{2}{*}{\textbf{Methods}} & \multicolumn{2}{c|}{\textbf{Kinetics700-QA}} & \multicolumn{2}{c|}{\textbf{STAR-bench-QA}} & \multicolumn{2}{c}{\textbf{FineDiving-QA}} \\
       & Accuracy & Score & Accuracy & Score & Accuracy & Score \\
       \midrule
      Video-LLaVA & 50.0 & 3.2 & 24.9 & 2.6 & 17.6 & 2.2 \\
      \base\ & 49.5 & 3.2 & 28.8 & 2.8 & 19.1 & 2.2\\
      \geminiBase\ & 41.9 & 2.9 & 22.3 & 2.6 & 16.3 & 2.1\\
       \rowcolor{blue} \ourMethod & 59.9 (+20\%) & 3.5 (+10\%) & 33.0 (+33\%) & 2.9 (+12\%) & 20.2 (+15\%) & 2.3 (+5\%) \\
      \bottomrule
    \end{tabular}
  }
  \vspace{-0.05in}
    \caption{\textbf{Adapted Dataset Performance.} Performance metrics on test sets of Kinetics700, FineDiving, and STAR-benchmark datasets via converting them to QA following~\citet{videoChatGpt}. \ourMethod\ shows significant improvement over Video-LLaVA and \base, showing the potential of \ourMethod\ for LVLM adaptation to new tasks.  \label{tab:dataset_adaptation}
}
\vspace{-0.05in}
\end{table}

\subsection{Quantitative Evaluation on Adapted Datasets}
\label{sec:res:adapt}
Besides improving general visual question-answering performance, \ourMethod can also adapt Large Vision-Language models to novel takes. To demonstrate this, we converted the test sets (\textit{not} included in training) of the source datasets -- Kinetics700, STAR-benchmark, and FineDiving. The results of these evaluations are reported in Tab.~\ref{tab:dataset_adaptation}.
Adapting LVLMs with easier-to-collect labels can be helpful in various applications, leading to a more versatile, multi-domain capable assistant.
When evaluating \ourMethod's impact on LVLM performance on the diverse source datasets, we found that it significantly improves model performance, particularly on complex tasks. For instance, on Kinetics700, known for its extensive action categories, \ourMethod\ enhanced Video-LLaVA's performance accuracy by an average of 20\% (as can be seen in Tab.~\ref{tab:dataset_adaptation}), showcasing its ability to develop generalized models adept across multiple domains. Interestingly, \base's performance did not improve compared to Video-LLaVA, and in some cases, even worsened, showing that one cannot directly utilize labeled datasets for LVLM adaptation.

Action Quality Assessment (AQA) is a complex video task requiring detailed action understanding, where \ourMethod\ significantly enhanced LVLM performance on the FineDiving dataset. Our results show a notable improvement from $17.6$ to $20.2$ in score prediction accuracy, highlighting \ourMethod's effectiveness in refining LVLM's temporal reasoning. 
However, \ourMethod allows LVLMs to not only rate a particular dive but also explain the rationale behind each assessment. This rationale is invaluable for many applications, effectively providing potential user feedback for improvement. 
This advancement enables novel applications, from sports coaching to automated feedback systems, by offering evaluations and constructive feedback.
The ability to interpret and improve action quality underscores the potential of \ourMethod, underscoring the potential of utilizing LVLMs as intelligent and informative visual assistants. For more, please see App. Sec.~\ref{sup:sec:aqa}.

\begin{table}[tb]
  \setlength\tabcolsep{3.4mm}
  \label{tab:ablation_adapted_datasets}
  \centering
  \adjustbox{width=\textwidth}{
    \begin{tabular}{l|cc|cc|cc}
      \toprule
      \multirow{2}{*}{\textbf{Ablations}} & \multicolumn{2}{c|}{\textbf{Kinetics700-QA}} & \multicolumn{2}{c|}{\textbf{STAR-bench-QA}} & \multicolumn{2}{c}{\textbf{FineDiving-QA}} \\
       & Accuracy & Score & Accuracy & Score & Accuracy & Score \\
       \midrule
        \ourMethod & \textbf{59.9} & \textbf{3.5} & \textbf{33.0} & \textbf{2.9} & \textbf{20.2} & \textbf{2.3} \\
        - Rationalization & 59.8 & \textbf{3.5} & 26.6 & 2.7 & 12.8 & 2.0 \\
        \ \ \  - Generation & 50.0 & 3.2 & 24.9 & 2.6 & 17.6 & 2.2 \\
      \bottomrule
    \end{tabular}
  }
  \vspace{-0.05in}
  \caption{\textbf{Ablations on Adapted Datasets.} Performance metrics on test sets of Kinetics700, STAR-benchmark, and FineDiving datasets. Label Rationalization impacts mostly the difficult datasets, such as FineDiving, whose initial Answer Generation yields are low.}
  \vspace{-0.05in}
\end{table}

\begin{table}[tb]
  \setlength\tabcolsep{1.7mm}
  \label{tab:ablation_vqa_datasets}
  \centering
  \adjustbox{width=\textwidth}{
  \begin{tabular}{l|cc|cc|cc|cc}
    \toprule
    \multirow{2}{*}{\textbf{Ablations}} & \multicolumn{2}{c|}{\textbf{MSVD-QA}} & \multicolumn{2}{c|}{\textbf{MSRVTT-QA}} & \multicolumn{2}{c|}{\textbf{TGIF-QA}} & \multicolumn{2}{c}{\textbf{ActivityNet-QA}} \\
     &   Accuracy & Score & Accuracy & Score & Accuracy & Score & Accuracy & Score \\
     \midrule

     \ourMethod &  \textbf{71.3} & \textbf{4.0} &\textbf{ 58.2} & \textbf{3.5} & {46.8} & {3.3} & {42.2} & {3.3} \\

    - Rationalization &  {70.6} & {3.9} & 57.5 & \textbf{3.5} & \textbf{47.7} & \textbf{3.4} & 42.2& 3.3 \\

    \ \ \  - Generation &  {69.7} & {3.9} & {57.4} & \textbf{3.5} & 46.5 & {3.3} & \textbf{43.2} & \textbf{3.4} \\
    \bottomrule
  \end{tabular} 
  }
  \vspace{-0.05in}
\caption{\textbf{Ablations on Zero-Shot Benchmarks}. In simpler benchmarks, Answer Generation proved more critical for zero-shot generalization than Label Rationalization.}
\vspace{-0.05in}
\end{table}

\subsection{Ablations}
In our ablation studies, we evaluated the impact of removing Label Rationalization and Answer Generation from \ourMethod, focusing on adapted datasets (Kinetics700, FineDiving, STAR-benchmark) and zero-shot benchmarks (MSVD-QA, MSRVTT-QA, TGIF-QA, ActivityNet-QA).

\paragraph{Adapted Datasets} For adapted datasets (Tab.~\ref{tab:ablation_adapted_datasets}), excluding Label Rationalization led to a significant performance drop in FineDiving, from $20.2$ to $12.8$ in accuracy, highlighting its critical role in complex reasoning tasks. This is likely due to the lack of conversion of any examples from the data. 
However, the removal of Answer Generation resulted in a more pronounced and uniform decline across all datasets. For example, Kinetics700's accuracy was reduced from $59.9$ to $50.0$, underscoring its foundational role in generating context-relevant responses. 

\paragraph{Zero-shot benchmarks}
In zero-shot benchmarks (Tab.~\ref{tab:ablation_vqa_datasets}), the removal of Label Rationalization had a mixed impact, slightly affecting MSVD-QA where accuracy decreased from $71.3$ to $70.6$. The elimination of Answer Generation consistently lowered performance, such as a decrease in MSRVTT-QA accuracy from $58.2$ to $57.4$.
ActivityNet-QA performance improved, probably because $100K$-Instruct utilizes ActivityNet for instruction tuning. Therefore, the introduction of additional videos decreases performance.

\vspace{-0.05in}
\section{Related Works}
\vspace{-0.05in}

\vspace{-0.05in}
\subsection{Large Vision-Language Models} 
\vspace{-0.05in}\label{sec:rw:lvlm}
Initial LVLMs, such as LLaVA~\citep{llava,llava2} and BLIP-2~\citep{BLIP2}, demonstrated the potential of merging image inputs with large language models. Methods like mPLUG-Owl~\citep{mplugowl} and Flamingo~\citep{flamingo} further allowed for multiple image inputs without architectural changes. \citet{videoChat} and \citet{llama_adapter} led the transition to video understanding, integrating video/image encoders and LLMs while training on small video instruction tuning datasets.
\citet{chatUniVi} introduced Chat-UniVi, a unified model employing dynamic visual tokens for both images and videos, optimizing visual token usage and higher frame count sampling.
LLaMA-VID~\citep{llama_vid} showed that the token count can be further reduced by pooling the tokens selectively via the text prompt using Q-Former.
Recently, Video-LLaVA~\citep{videoLlava} used modality-specific encoders for video and image inputs to leverage LanguageBind encoders as they are constructively aligned during pretraining and utilized a shared projection.

\citet{videoChatGpt} expanded the field with the first large video instruction tuning dataset, VideoInstruct-$100K$. 
VideoInstruct-$100K$ was generated from ActivityNet~\citep{ActivityNet} by prompting chatGPT with the video captions, generating question-answer pairs. 
While driving much of the performance improvement in the field~\citep{chatUniVi,VaQuitA,videoLlava, llama_vid}, upon examination of VideoInstruct-$100K$, it is evident that it suffers from quality issues. 
The questions often degrade into de facto prompts for a video caption (see Fig.~\ref{sup:fig:video_instruction_quality}) and rarely require many spatiotemporal capabilities, which may limit LVLM performance.

\vspace{-0.05in}
\subsection{Large Language Models and Self-Training}\label{sec:rw:llm}
\vspace{-0.05in}
The advent of GPT~\citep{GPT1, GPT2} marked significant milestones in natural language processing, showcasing LLMs' power in understanding and generating human-like text. Open-source LLMs like LLaMA~\citep{llama, llama2} and their instruction-tuned variants like Alpaca and Vicuna~\citep{Alpaca, Vicuna} further tailored these models for nuanced human-AI interactions. 
However, even LLMs have found it challenging to scale annotated datasets for training, prompting work on self-training and self-improvement~\citep{SelfTraining, SelfImprove, SSCoT1, SSCoT2, VSTaR}. In this line of work, LLMs cycle between instruction-tuning data generation and instruction tuning, iteratively improving LLM performance over cycles. For instance,~\citet{STaR} introduced the 
the Self-Taught Reasoners method, used rationalization to generate chain-of-thought (CoT) reasoning, filtering poor rationalizations to retain correctly answered questions. Other self-training approaches include expectation-maximization-based approaches~\citep{SelfTraining}, which alternate between data generation and improvement between training cycles. Alternatively, majority voting has also been utilized to generate answers and rationale for unlabeled questions~\citep{SelfImprove}. These methods show the effectiveness of iterative self-training. In our work, we aim to introduce a weakly supervised self-training approach for video instruction tuning, leveraging video supervision that is often easier to collect and exists in many large and diverse datasets.

\vspace{-0.05in}
\section{Conclusions}
\vspace{-0.05in}
In conclusion, \underline{Video} \underline{S}elf-\underline{Ta}ught \underline{R}easoners (\ourMethod) presents a novel approach to enhance Large Vision-Language Models (LVLMs) by enabling the use of diverse labeled video datasets for visual instruction tuning. This method addresses critical data diversity and quality challenges, leading to significant performance improvements across various video understanding tasks. Our experiments demonstrate \ourMethod's effectiveness in both source dataset adaptation and zero-shot generalization, showcasing its potential in advancing LVLM capabilities for complex video content.

The promising results of \ourMethod\ open new research avenues, particularly in expanding LVLM knowledge bases using readily available image and video datasets. Future work could explore advanced self-training techniques and integration with emerging LVLM architectures, focusing on long-form video understanding to further boost LVLM understanding.
Additional work is also needed to reduce hallucinations, perhaps by using grounded VLMs as auxiliary input. 

\vspace{0.12in}

\noindent\textbf{Acknowledgements.} We thank Google Research for providing financial support through a Stanford HAI–Google collaboration. This work was partially supported by the National Science Foundation under Grant No. 2026498. We thank the Knight-Hennessy Scholars Foundation for funding OZ.

\vspace{-0.05in}
\section{Limitations}
\vspace{-0.05in}
While \ourMethod\ introduces a novel approach to visual instruction tuning, it is not without its limitations. Firstly, the methodology can be computationally intensive due to the cycling of both generating and rationalizing question-answer and instruction tuning. Secondly,   the assumption that all labels necessitate a rationale may not always hold true. Certain labels might be straightforward enough not to require elaborate rationalization, potentially leading to unnecessary computational overhead. Lastly, hallucinations, especially in label rationalization can be further reduced by perhaps implementing additional verifiers.

\bibliography{iclr2024_conference}
\bibliographystyle{iclr2024_conference}

\appendix
\newpage

\section{Appendix}

In Sec.~\ref{sup:sec:implementation_details}, we provide additional implementation details and compute used in developing \ourMethod. 
In Sec.~\ref{sup:sec:aqa}, we introduce explainable action quality assessment and provide good and bad examples of \ourMethod\ on the FineDiving test dataset. 
Finally, we provide additional qualitative Answer Generation and Label Rationalization examples in Sec.~\ref{sup:sec:answer_gen} and~\ref{sup:sec:rationalize}.

\subsection{Implementation Details}\label{sup:sec:implementation_details}

\ourMethod\ utilized the Video-LLaVA model, which integrated the Vicuna-7B v$1.5$ for language processing and ViT-L/14 video and image encoders from LanguageBind for visual encoding. The system's tokenizer, adapted from LLaMA, has a vocabulary size of around $32,000$ classes and a dimensionality of $4096$.
Two cycles of \ourMethod\ were executed, each initialed with the pre-trained Video-LLaVA model (before instruction tuning). The training data was augmented by incorporating VideoInstruct-$100$K and LLaVA v$1.5$'s visual instruction datasets.

Four clusters of $10$ NVIDIA Titan RTX GPUs were employed for $64$ hours. The structured prompts for these tasks were as follows:

\begin{itemize}
    \item \textbf{Answer Generation}
    \begin{quote}
    \texttt{Question: \{Q\}.\\
    Can you explain step-by-step how one can arrive at this\\ conclusion?}
    \end{quote}
    \item \textbf{Label Rationalization}
\begin{quote}
    \texttt{Question: \{Q\}\\
    Answer: \{L\}.\\
    Can you explain step-by-step how one can arrive at this \\ conclusion?}
\end{quote}

\end{itemize}

 \begin{figure*}[!tb]
  \centering
    \resizebox{\linewidth}{!}{
    \includegraphics{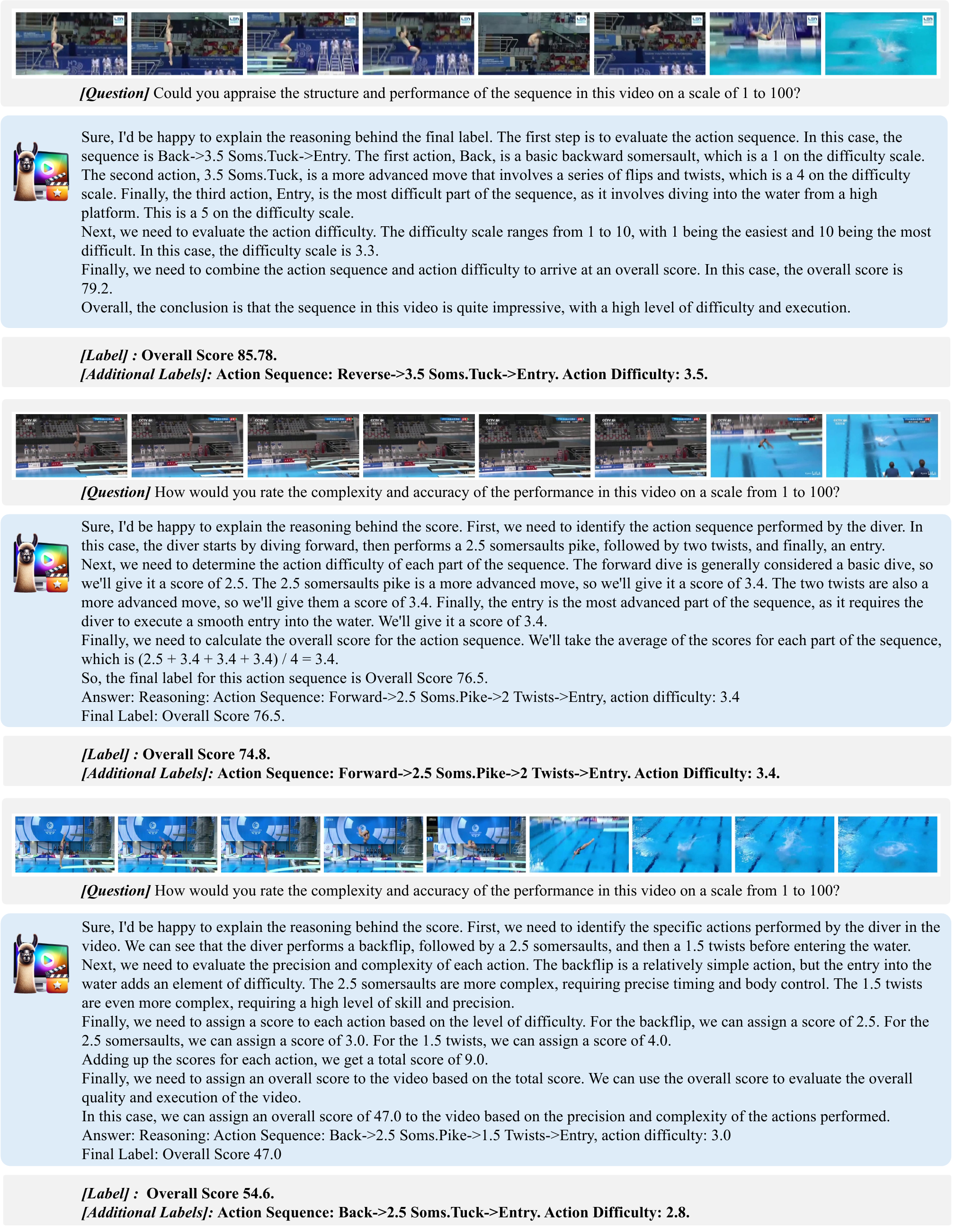}
    }
  \caption{
  \textbf{Action Quality Assessments by \ourMethod\ on the FineDiving Test Set.} 
  Different diving sequences with corresponding \ourMethod\ evaluations, from a high score of $85.78$ for complex sequences (top) to $74.8$ for intermediate (middle), and a lower score of $54.6$ for basic sequences (bottom), showcasing \ourMethod's proficiency in scoring dives with varying degrees of difficulty and execution quality.
  }
\label{fig:goodaqa}
\end{figure*}

These prompts guided the model in producing detailed answers and rationalizations, enhancing the depth and utility of the generated instruction-tuning dataset.
Answer correctness was evaluated using template matching with Levenshtein Distance-based~\cite{fuz1} fuzzy logic~\cite{fuz2}, considering an answer correct if all keywords from the label were present in the generated response with a minimum similarity score of $80$\%. For example, if in Kinetics the action label is `eating apple pie', we would only consider a generated answer correct of `eating', `apple', `pie' all appeared with a similarity score of $80$.

\subsection{Explainable Action Quality Assessment}\label{sup:sec:aqa}

Explainable Action Quality Assessment (AQA) is critical for detailed analysis of performances in precision sports, such as competitive diving, where execution and complexity significantly impact scores. Unlike previous AQA works, which only provide a score~\cite{FineDiving, AQA_1, AQA_2, AQA_logo} \ourMethod\ not only generates scores but also offers detailed justifications akin to expert analysis.

Fig.~\ref{fig:goodaqa} illustrates instances where \ourMethod's predictions align closely with established scoring criteria in diving. For example, the model breaks down a dive sequence, Reverse$\rightarrow$$3.5$ Soms.Tuck$\rightarrow$Entry, with an assigned difficulty of $3.5$, into its components. It then logically assigns scores to each element, such as $1$ for more straightforward maneuvers and $4$ for more complex twists, culminating in an overall difficulty score of $3.3$, close to the GT $3.5$. The final predicted score was $79.2$, close to the GT score of $85.78$.

 \begin{figure*}[!tb]
  \centering
    \resizebox{\linewidth}{!}{
    \includegraphics{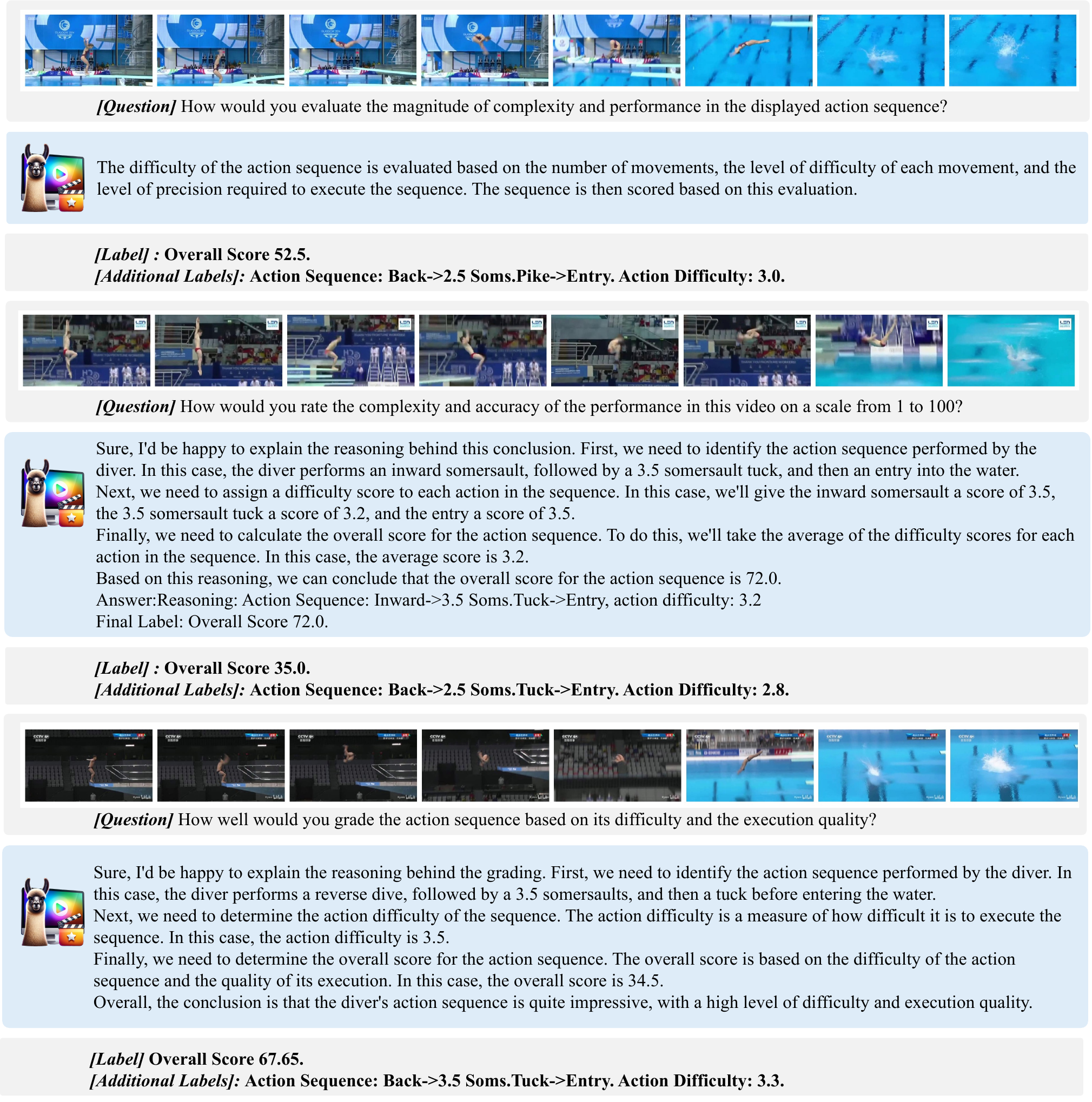}
    }
    \vspace{-0.17in}
  \caption{
  \textbf{Challenges in Action Quality Assessment by \ourMethod\ on the FineDiving Test Set.} Illustrating scenarios where \ourMethod\ diverges from optimal scoring: a complete absence of prediction (top), an overestimation of a dive's complexity and execution (middle), and underestimation (bottom).
  \vspace{-0.07in}
  }
\label{fig:bad_aqa}
\end{figure*}

 \begin{figure*}[!t]
  \centering
    \resizebox{\linewidth}{!}{
    \includegraphics{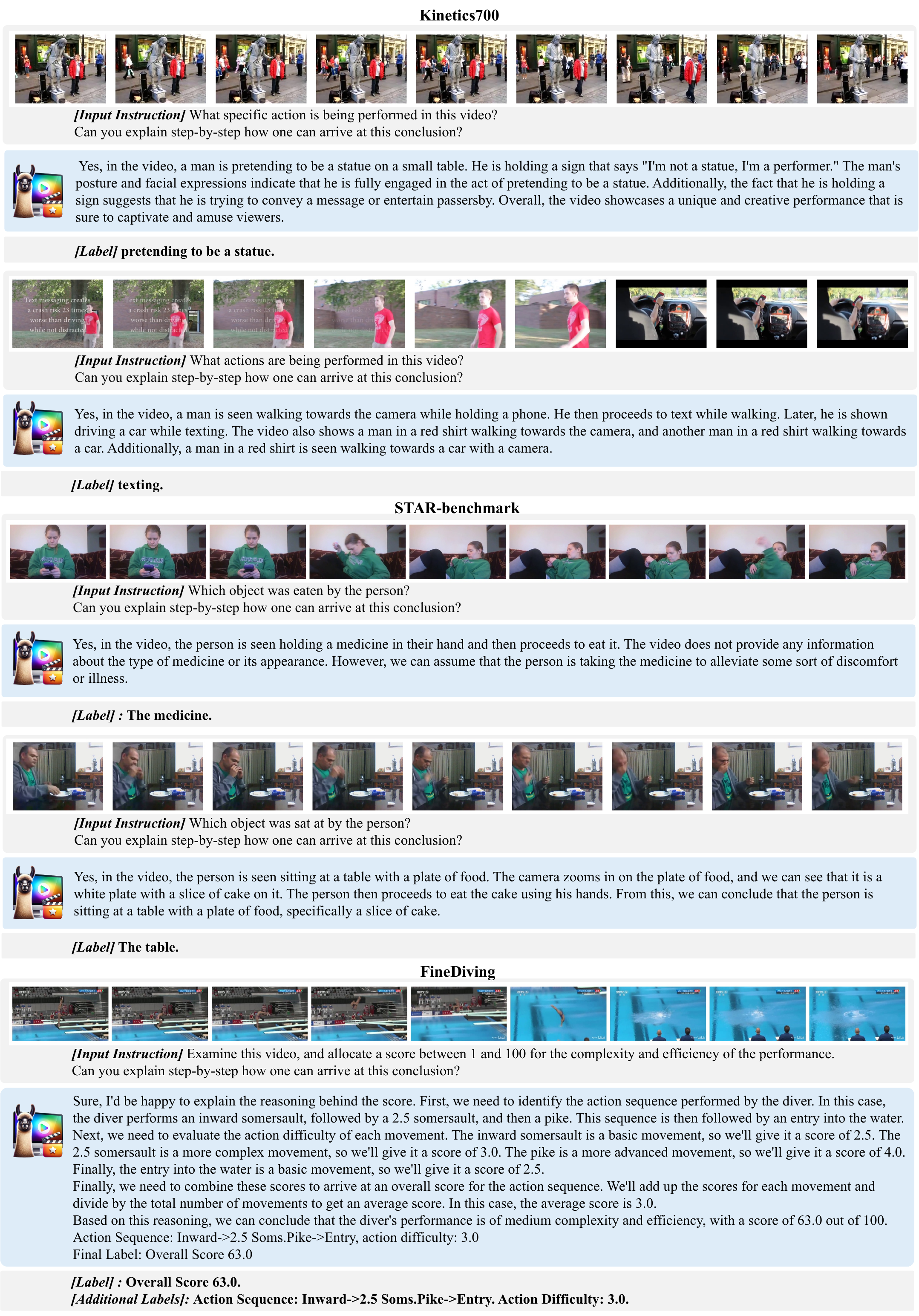}
    }
  \caption{
  \textbf{Answer Generation Across Datasets by \ourMethod.} Interpreting actions on Kinetics700 (top), detailing action sequences in STAR-benchmark (middle), and evaluating diving action quality in FineDiving (bottom).
  }\vspace{-0.15in}
\label{fig:answer_gen_examples}
\end{figure*}

\ourMethod's proficiency extends to dives with varying levels of performance. It can discern relatively complex dives with ground-truth scores of  $74.8$ and $54.6$, which \ourMethod\ scored as $76.5$ and $47.0$, respectively. In both cases, the model breaks down the actions into sub-actions and rates them in terms of difficulty and execution. While it manages to rate the dives themselves well, in one instance, the model erroneously calculated the average for $2.5,3.4,3.4,3.4$ as $3.4$.

Challenges in maintaining consistent AQA accuracy are depicted in Fig.~\ref{fig:bad_aqa}, showcasing instances of either the model not following instructions or estimating the score incorrectly. In Fig.~\ref{fig:bad_aqa}, top, the model did not produce a score and resorted to explaining how a score might be derived. 
The model occasionally produces an inaccurate score, and either grossly over (Fig.~\ref{fig:bad_aqa}, middle) or under (Fig.~\ref{fig:bad_aqa}, bottom) estimates the score. For example, it attributed a score of $72.0$ to a sequence deemed to have a lower difficulty level of $35.0$. These discrepancies underscore the necessity for ongoing improvements in the model's grasp of nuanced scoring criteria to enhance reliability in AQA predictions.

In summary, \ourMethod\ enhances Action Quality Assessment (AQA) by supplementing scores with detailed rationales, an advancement over traditional AQA approaches that only provide numerical scores. Although it effectively dissects the components of diving performances, indicating both complexity and execution as seen in Fig.~\ref{fig:goodaqa}, it faces challenges in maintaining consistent accuracy, particularly in aligning scores with established benchmarks, as evidenced in Fig.~\ref{fig:bad_aqa}. These instances highlight the need for a deeper understanding of each movement's difficulty and execution quality to ensure the model's scoring aligns with professional judging standards. Additionally, they emphasize the ongoing requirement to improve the model's accuracy and ground its rationalizations in the verifiable aspects of the performances it evaluates.

\subsection{Qualitative Analysis of Answer Generation}\label{sup:sec:answer_gen}
Fig.~\ref{fig:answer_gen_examples} presents our model's capabilities in generating answers and rationalizing actions across varied video contexts, demonstrating its adeptness in interpreting complex scenes.

\paragraph{Kinetics700.} 
In the first example from Kinetics700 (Fig.~\ref{fig:answer_gen_examples}, Kinetics700, top), the model effectively identifies a man's act of pretending to be a statue and further discerns the performance's subtle aspects, such as the engagement level and the humor conveyed through a sign. 
In another Kinetics700 example (Fig.~\ref{fig:answer_gen_examples}, Kinetics700, bottom), the model processes a scene with multiple concurrent actions. \ourMethod\ first correctly identifies the man in the red shirt talking towards the camera while holding a phone and texting. It correctly identifies the next scene, where another man is texting while driving. 
This precision in temporally locating different actions in the videos is invaluable for visual instruction tuning and could potentially enhance models'
 The capability to analyze scenes with multiple focal points is an essential feature for comprehensive video understanding.

 \begin{figure*}[t]
  \centering
    \resizebox{\linewidth}{!}{
    \includegraphics{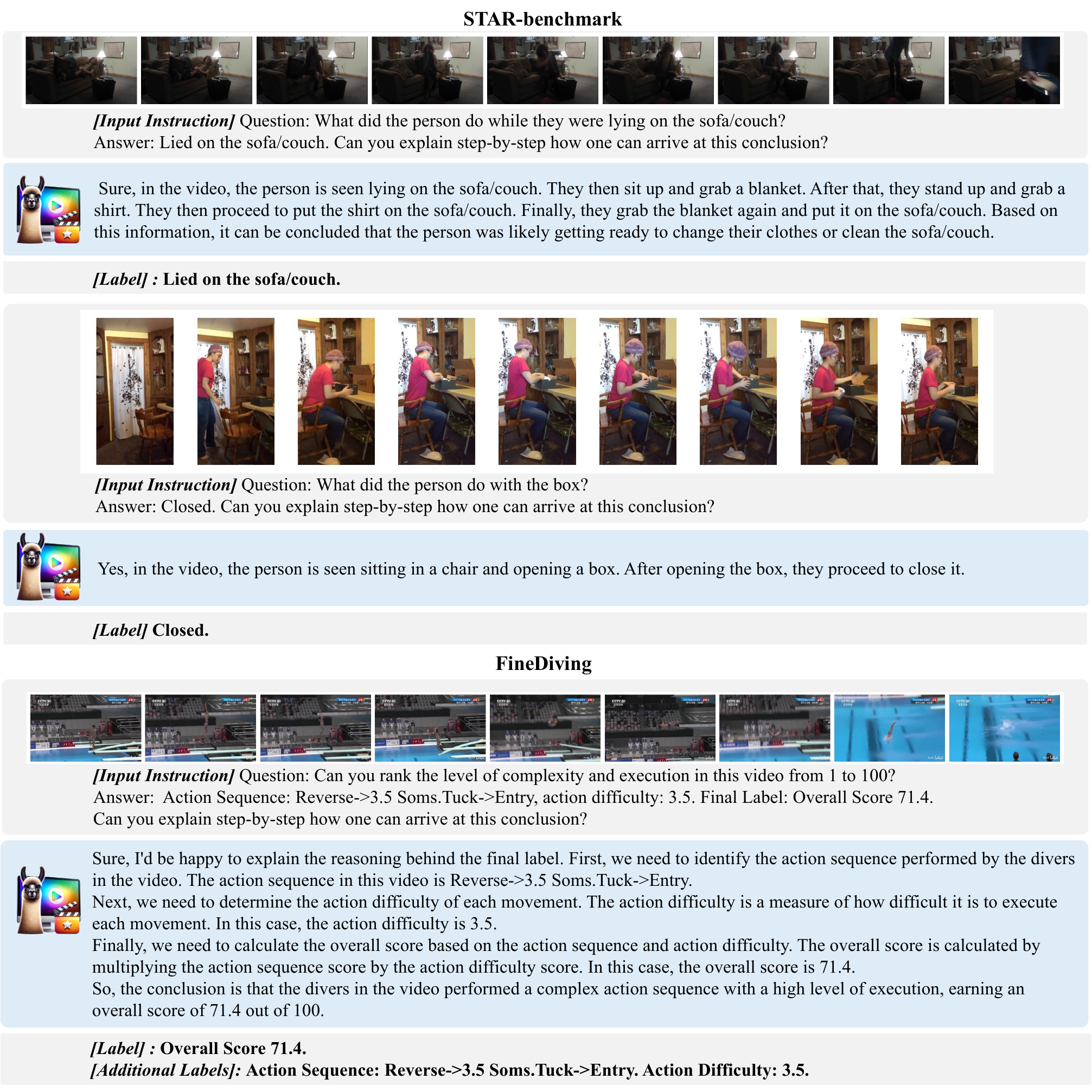}
    }
  \caption{
  \textbf{Successful Label Rationalizations by \ourMethod.} The model accurately infers a preparatory behavior from a person's interaction with household items (top), correctly interprets box-handling actions (middle), and provides a nuanced breakdown of a complex diving sequence, assigning an overall score of $71.4$ (bottom), exemplifying precise action understanding on FineDiving and STAR-benchmark.
  \vspace{-0.1in}
  }
\label{fig:label_rat_good}
\end{figure*}

 \begin{figure*}[!ht]
  \centering
    \resizebox{\linewidth}{!}{
    \includegraphics{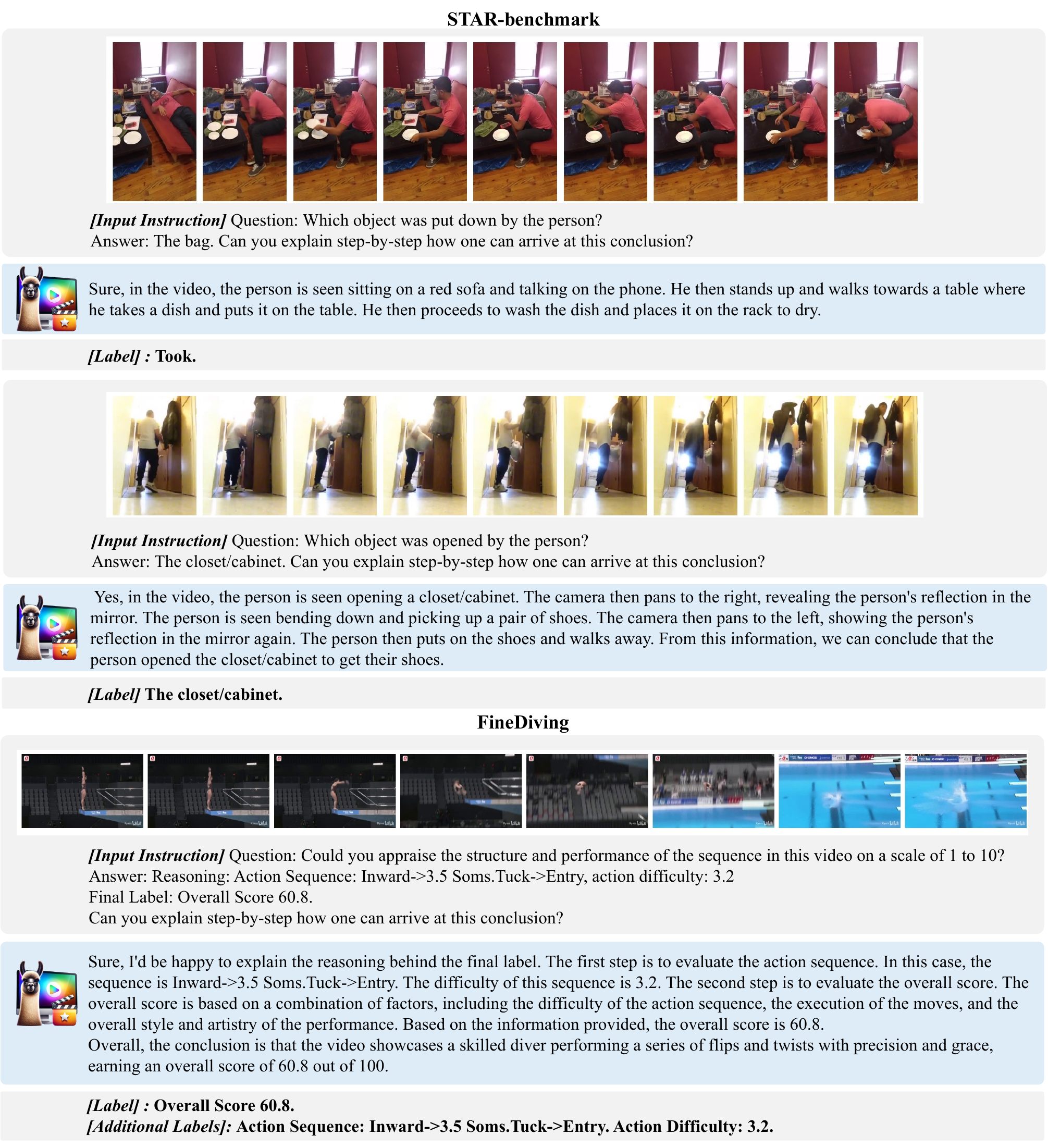}
    }
  \caption{
  \textbf{Challenges in Label Rationalization by \ourMethod.} Instances of rationalization errors are shown: an incorrect inference about dishwashing (top) and fabricated details about closet interactions (middle). A rationalization for a diving sequence (bottom) is accurate but demonstrates the model's vulnerability to hallucinations in complex action sequences within FineDiving and STAR-benchmark datasets.
  }
\label{fig:label_rat_bad}
\end{figure*}

\paragraph{STAR-benchmark.} In the first STAR-benchmark example (Fig.~\ref{fig:answer_gen_examples}, STAR- benchmark, top), the model uses inferential reasoning to deduce the purpose behind a person consuming medicine despite the absence of explicit details about the medicine. This instance showcases the model's ability to apply logical assumptions to fill in informational gaps, a valuable trait for interpreting actions without fully detailed context. In the next example (Fig.~\ref{fig:answer_gen_examples}, STAR-benchmark, bottom), \ourMethod\ identified additional details, such as the person sitting at the table eating cake.

\paragraph{FineDiving.} in Fig.~\ref{fig:answer_gen_examples}, FineDiving, \ourMethod's approach to evaluating a diving sequence in FineDiving illustrates its proficiency in detailed performance analysis. 
By deconstructing the dive into individual elements and assessing each for difficulty, the model mirrors the evaluative processes of human judges, providing a comprehensive performance score. 
This depth of analysis, which includes a critique of the dive's complexity and execution, underscores the model's utility in contexts requiring nuanced assessment, such as athletic performance evaluation.

These examples from Fig.~\ref{fig:answer_gen_examples} show how the proposed Answer Generation is capable of creating valuable and informative video question-answer pairs that can be utilized in instruction tuning, highlighting its potential applicability in various domains that demand a deep understanding of video scenes.

 \begin{figure*}[!t]
  \centering
    \resizebox{\linewidth}{!}{
    \includegraphics{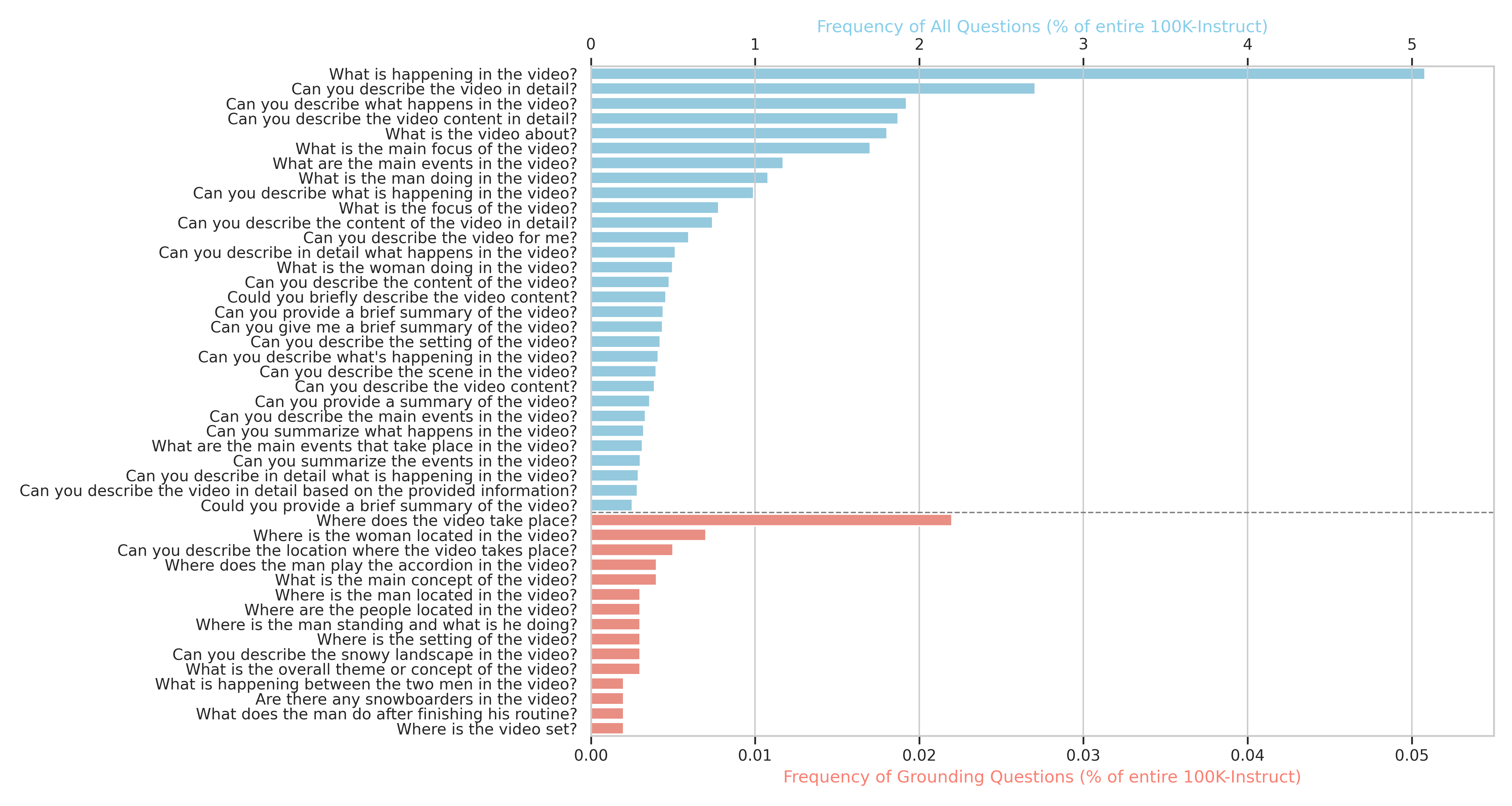}
    }
  \caption{
  \textbf{VideoInstruct-$100K$ Qualitative Evaluation.} Evaluation of VideoInstruct-$100K$'s question distribution. A wide gap can be seen between `grounding' questions, which contains `where/when' (bottom) and the top-50 most common questions (top). when analyzing the top-50 most common questions (top), it is clear they are all prompting for video captions. 
  }
\label{sup:fig:video_instruction_quality}
\end{figure*}

\subsection{Qualitative Examples of Label Rationalizations}\label{sup:sec:rationalize}

Label rationalization in \ourMethod\ is initiated when direct answer generation by the Large Multimodal Model (LMM) fails. Although this method proves beneficial in certain situations, it occasionally leads to the generation of hallucinated content—incorrect details or inferences not supported by the video.

\paragraph{STAR-benchmark. } Illustrated in Fig.~\ref{fig:label_rat_good}, effective label rationalizations provide added context, enriching the model's responses. A notable example from the STAR benchmark demonstrates the model's capacity to build upon a basic action, like lying on a couch, by inferring additional activities, such as donning a shirt and tidying up, hinting at the individual's subsequent actions. This example illustrates \ourMethod's ability to navigate ambiguous labels and furnish more nuanced, informative responses, crucial for Visual Instruction Tuning.

However, Label Rationalization is also more prone to hallucinations. Fig.~\ref{fig:label_rat_bad} exposes the model's tendency for hallucinations, mainly where it introduces actions and details not evidenced in the video. In Fig.~\ref{fig:label_rat_bad}, STAR-benchmark, top; the LMM hallucinated that after taking the dish, the man washed it and placed it on a dry rack, which did not occur in the video. In Fig.~\ref{fig:label_rat_bad}, STAR-benchmark, bottom; the model hallucinated that the camera panned to the right, that it could see the reflection of the man in the mirror, and that he took out shoes from the closet — none of which occur in the video.

\paragraph{FineDiving.} Label rationalization proved especially useful in challenging, domain-expert datasets such as the Olympic diving scoring dataset - FineDiving. 
Answer Generation initially had zero yield, and Label Rationalization allowed the model to start learning about this challenging task. 
For example, in Olympic diving events, to get an overall score for the dive, one removes the top and bottom $2$ scores (out of a total of $7$), then multiplies this score with the dive difficulty. In Fig.~\ref{fig:label_rat_good}, FineDiving, the LMM correctly deduced that the execution score is multiplied by the difficulty. 
With no supervision, the model correctly deduced the rules, deducing that the final execution score is obtained by multiplying the execution score and the action difficulty.

Fig.~\ref{fig:label_rat_bad}, FineDiving, Label Rationalization failed to generate an answer with sufficient depth. Rather than providing additional insights, the model resorted to re-iterating the labels.

In summary, Label Rationalization produces shorter, less informative answers than Answer Generation and is more prone to hallucinations. It is primarily utilized so complex datasets, such as FineDiving, can be learned, especially in cases with initial zero yield.

\subsection{Evaluation of Video Instruction Tuning Datasets}

We performed a qualitative evaluation of VideoInstruct-$100K$ and found that the broad majority of the questions essentially prompt the Large Vision-Language model for video captions - see Fig.~\ref{sup:fig:video_instruction_quality}. As can be seen, $\sim75\%$ of VideoInstruct-$100K$'s questions are of this type.

\end{document}